\documentclass[10pt,twocolumn,letterpaper]{article}

\usepackage{cvpr}

\newcommand{\Description}[1]{}

\usepackage{booktabs}
\usepackage{tikz}
\usepackage[ruled]{algorithm2e}

\SetAlFnt{\small}
\SetAlCapFnt{\small}
\SetAlCapNameFnt{\small}
\SetAlCapHSkip{0pt}

\definecolor{cvprblue}{rgb}{0.21,0.49,0.74}
\usepackage[breaklinks,colorlinks,allcolors=cvprblue]{hyperref}

\def\paperID{*****}
\def\confName{CVPR}
\def\confYear{2026}

\title{StreamChar: Long-Horizon Streaming Character Audio-Video Generation with Decoupled Orchestration}

\author{Linrui Tian\thanks{Equal contribution.}\and Qi Wang\footnotemark[1]\and Bang Zhang \\
Tongyi Lab, Alibaba Group \\
{\tt\small tianlinrui.tlr@alibaba-inc.com \quad wilson.wq@alibaba-inc.com \quad zhangbang.zb@alibaba-inc.com} \\
{\tt\small Project page: \url{https://humanaigc.github.io/StreamChar_page}}}

\begin{document}
\maketitle

\begin{abstract}
    Real-time streaming joint audio-video generation for character animation requires a generator to speak the requested transcript, maintain visual identity across chunks, and run within a strict playback budget. These requirements are difficult to satisfy simultaneously: chunk-wise autoregressive generation can accumulate transcript-audio misalignment and visual drift, while the few-step distillation needed for low latency often degrades spatial diversity and temporal quality. We present StreamChar, a streaming framework that separates long-horizon orchestration from short-window audio-video denoising. An LLM-based orchestrator uses the transcript and historical context to produce frame-aligned audio conditions, and a joint audio-video DiT performs local bidirectional denoising with reference and motion-frame conditioning. For efficient deployment, we use a two-stage distillation pipeline that first compresses the sampler and then fine-tunes the student under online chunk rollouts. A progress-aware pointer aligns partial transcripts with generated audio during rollout training, and a sink-chunk memory provides a persistent visual anchor for reducing long-horizon drift. Experiments on short-clip and long-horizon protocols show that StreamChar runs in real time on a single H100 GPU and provides a favorable system-level trade-off among transcript fidelity, audio-visual synchronization, visual quality, and streaming stability compared with recent joint and audio-driven baselines.
\end{abstract}

\section{Introduction}

Real-time streaming joint audio--video generation for characters from text
is a challenging problem at the intersection of multimodal learning,
efficient inference, and interactive systems. Recent advances in latent
diffusion~\cite{rombach2022high} and diffusion transformers
(DiT)~\cite{peebles2023scalable} have enabled high-quality short-clip
generation, and efforts to unify text, audio, and video within a single
backbone~\cite{low2025ovitwinbackbonecrossmodal, hacohen2026ltx2efficientjointaudiovisual, seedance2026seedance20advancingvideo} have pushed joint modeling further.
 However, the transition from \emph{generating clips} to \emph{streaming
minutes-long content interactively} surfaces two tightly coupled but
fundamentally distinct difficulties: \textbf{long-horizon coherence} and
\textbf{interactive inference speed}.

The first difficulty, long-horizon coherence, arises from the need to generate
content chunk by chunk over extended durations. A critical requirement, often
under-specified in recent multimodal DiT systems, is maintaining strict
correspondence between the cumulatively generated audio and the global input
transcript across multiple chunks. In chunk-wise autoregressive settings,
local decoding decisions can drift from the textual plan, leading to omitted
content, repetitions, or semantic misalignment with the source script. Beyond
this transcript--audio fidelity, the model must simultaneously preserve
speaker identity and visual appearance across segment boundaries, and maintain
frame-accurate lip--phoneme alignment as the sequence lengthens.
In monolithic multimodal DiT designs, the
same backbone shoulders all three responsibilities -- semantic understanding,
cross-chunk memory, and local spatiotemporal denoising. This creates competition
for model capacity: errors in global context propagate directly into local generation,
manifesting as semantic drift, identity shifts, and degraded synchronization
after the first few chunks. The problem is compounded by autoregressive
rollouts, where each chunk's errors become the conditioning context for the
next, creating a feedback loop that rapidly diverges from the intended content.

The second difficulty, \textbf{interactive inference speed}, is equally critical.
Real-time streaming requires each chunk to be generated faster than its
playback duration, yet diffusion models typically need tens to hundreds of
denoising steps for quality. Aggressive step reduction via distillation is
necessary but introduces \emph{distillation-induced mode collapse}: the student
model, deprived of iterative refinement, collapses to stereotyped spatial
behaviors and reduced diversity. Moreover, when deployed autoregressively for
chunk-by-chunk streaming, errors accumulate across chunks, causing progressive
\emph{video drifting} that further degrades long-horizon quality.

Critically, these two challenges are not independent. The quality degradation
from aggressive distillation amplifies the error accumulation of autoregressive
generation, while long-horizon instability makes it harder to train a robust
few-step student. To address them jointly, we propose \textbf{StreamChar}, which
coordinates design at two levels: an \emph{architecture} that distributes
responsibility across specialized components, and an \emph{optimization strategy}
that sequentially resolves step reduction and rollout consistency.

For \textbf{long-horizon coherence}, we adopt a decoupled LLM + DiT architecture.
An LLM orchestrator reads the transcript and historical context, producing
compact frame-aligned audio conditions $\mathbf{c}_a$ that specify what should be
acoustically expressed in the current chunk. The DiT backbone then focuses on
short-window joint audio--video denoising with full bidirectional attention,
rather than carrying the entire burden of script tracking and local synthesis.
Cross-chunk continuity is maintained through explicit \textbf{motion frame
conditioning}, where previously generated chunks provide temporal context.
For \textbf{interactive inference speed}, we design a two-stage decoupled
distillation pipeline. Stage~I applies distribution matching distillation (DMD)
to compress the teacher into a few-step student, focusing purely on single-chunk
generation quality. Stage~II then fine-tunes this student with \emph{online
rollout simulation}, where the model generates multiple consecutive chunks
autoregressively and is optimized on its own outputs.

Two key designs enable stable Stage~II training: (1) a \textbf{progress-aware
pointer (PAP)} that predicts the transcript endpoint for each generated chunk,
aligning partial transcripts with generated audio; and (2) a \textbf{sink-frame
mechanism} where the first chunk serves as a persistent long-range anchor
attended by all subsequent chunks, suppressing video drifting in long rollouts.
This two-stage decoupling reduces the gradient interference observed when step
compression and rollout consistency are optimized simultaneously.

In summary, this work makes the following contributions:
\begin{itemize}
  \item A \textbf{decoupled LLM orchestrator + short-window DiT} architecture
    that addresses long-horizon coherence by offloading global semantics from
    the denoising backbone, with motion frame conditioning for cross-chunk
    continuity.
  \item A \textbf{two-stage distillation recipe} that decouples step
    compression (Stage~I) from rollout consistency training (Stage~II),
    with a progress-aware pointer for transcript alignment and a sink-frame
    mechanism for suppressing long-horizon quality drift.
  \item Quantitative and qualitative evaluation demonstrating that StreamChar
    supports long-horizon real-time streaming audio-video generation for
    characters on a single GPU, while remaining competitive with recent
    streaming and non-streaming methods.
\end{itemize}

\section{Related Work}

\paragraph{Diffusion models for audio-video generation.}
Denoising diffusion probabilistic models~\cite{ho2020denoising} and latent diffusion~\cite{rombach2022high}, particularly Diffusion Transformers (DiT)~\cite{peebles2023scalable}, form the backbone of modern generative media. Recent works unify text, audio, and visual tokens in monolithic DiTs for joint generation~\cite{openmoss_mova_2026, hacohen2026ltx2efficientjointaudiovisual, low2025ovitwinbackbonecrossmodal}. While effective for single short clip, these monolithic designs face challenges when scaled to long-form streaming scenarios. 
The shared backbone must simultaneously handle semantic understanding, cross-segment memory, and local spatiotemporal denoising, leading to capacity competition and error propagation across chunks.
Moreover, recent real-time streaming methods~\cite{li2026halloliverealtimestreamingjoint} are typically confined to short temporal windows (few seconds).

\paragraph{LLMs as planners and conditioners.}
Large language models~\cite{vaswani2017attention} have demonstrated remarkable capabilities in high-level semantic understanding and structured reasoning, leading to their widespread adoption as central planners in generative AI. 
In visual synthesis, LLMs are increasingly employed to decompose complex prompts into structured layout specifications~\cite{liu2025javisdit, wu2025qwenimagetechnicalreport} or storyboards~\cite{jiang2025omnihuman15instillingactivemind} that guide downstream diffusion models. 
Similarly, in the audio domain, recent works~\cite{peng2025vibevoicetechnicalreport, cosyvoice} leverage LLMs to bridge the gap between textual semantics and acoustic realization, utilizing them to generate control signals or intermediate representations for speech synthesis. 
These approaches highlight an emerging paradigm where LLMs handle long-range contextual consistency and global script semantics, while generative backbones focus on local fidelity.

\paragraph{Audio-driven video generation.}
Audio-driven video generation has evolved from offline quality-oriented methods to real-time streaming systems. Early works such as EMO~\cite{tian2024emo} and Wan2.2-S2V~\cite{wans2v} demonstrated high-fidelity portrait animation using diffusion models, but require dozens of denoising steps unsuitable for interactive applications. Recent streaming approaches~\cite{zeng2026lpm10videobasedcharacter,shen2026soulxflashtalkrealtimeinfinitestreaming,yu2026soulxflashheadoracleguidedgenerationinfinite,huang2026liveavatarstreamingrealtime} achieve sub-second latency through knowledge distillation, yet operate in the audio-driven paradigm where video is synthesized from a given waveform. This simplifies synchronization and makes these systems strong references for video quality and latency, but it does not address the harder text-to-audio-video setting where speech content and visual motion must be generated together. StreamChar targets this joint setting and therefore requires explicit cross-modal coordination through decoupled LLM orchestration and short-window bidirectional DiT denoising.

\paragraph{Knowledge distillation for efficient diffusion.}
Reducing the inference cost of diffusion models is critical for real-time
applications. Progressive distillation, consistency models, and distribution
matching distillation (DMD)~\cite{yin2024onestepdiffusiondistributionmatching}
have been proposed to compress multi-step samplers into few-step generators.
However, when these distilled models are deployed in autoregressive or
chunk-wise streaming settings, they encounter two intertwined failure modes:
\emph{distillation-induced mode collapse}, where the student converges to a
narrow set of spatial behaviors, and \emph{error accumulation}, where
imperfections in early chunks propagate forward and compound over time.
Recent efforts have attempted to address these issues in isolation~\cite{huang2025selfforcingbridgingtraintest,chen2024diffusionforcingnexttokenprediction}.
Decoupled DMD~\cite{liu2025decoupleddmdcfgaugmentation} improves training stability for
long-horizon image generation by mitigating mode collapse, but remains
limited to static images without considering temporal dynamics or cross-chunk
error propagation in video. Self-forcing and rolling rollout
methods~\cite{huang2025selfforcingbridgingtraintest,cui2025selfforcingminutescalehighqualityvideo} explicitly train
on sequential generation to reduce error accumulation, but assume access to
a pre-trained teacher with stable single-step distillation, overlooking the
interaction between mode collapse and rollout instability. Prior work has
largely treated step reduction and rollout consistency as separate problems\cite{nie2026transitionmatchingdistillationfast}.
We show that these challenges are deeply coupled in streaming scenarios
and propose a two-stage distillation strategy that sequentially resolves them:
Stage~I compresses the sampler via DMD on single chunks, and Stage~II performs
online rollout fine-tuning with a sink-frame mechanism to suppress video
drifting. This decoupling is central to achieving both interactive speed and
long-horizon stability.

\section{Method}
\label{sec:methodoverview}

\subsection{Overall architecture}

\begin{figure*}[t]
  \centering
  \includegraphics[width=0.7\textwidth]{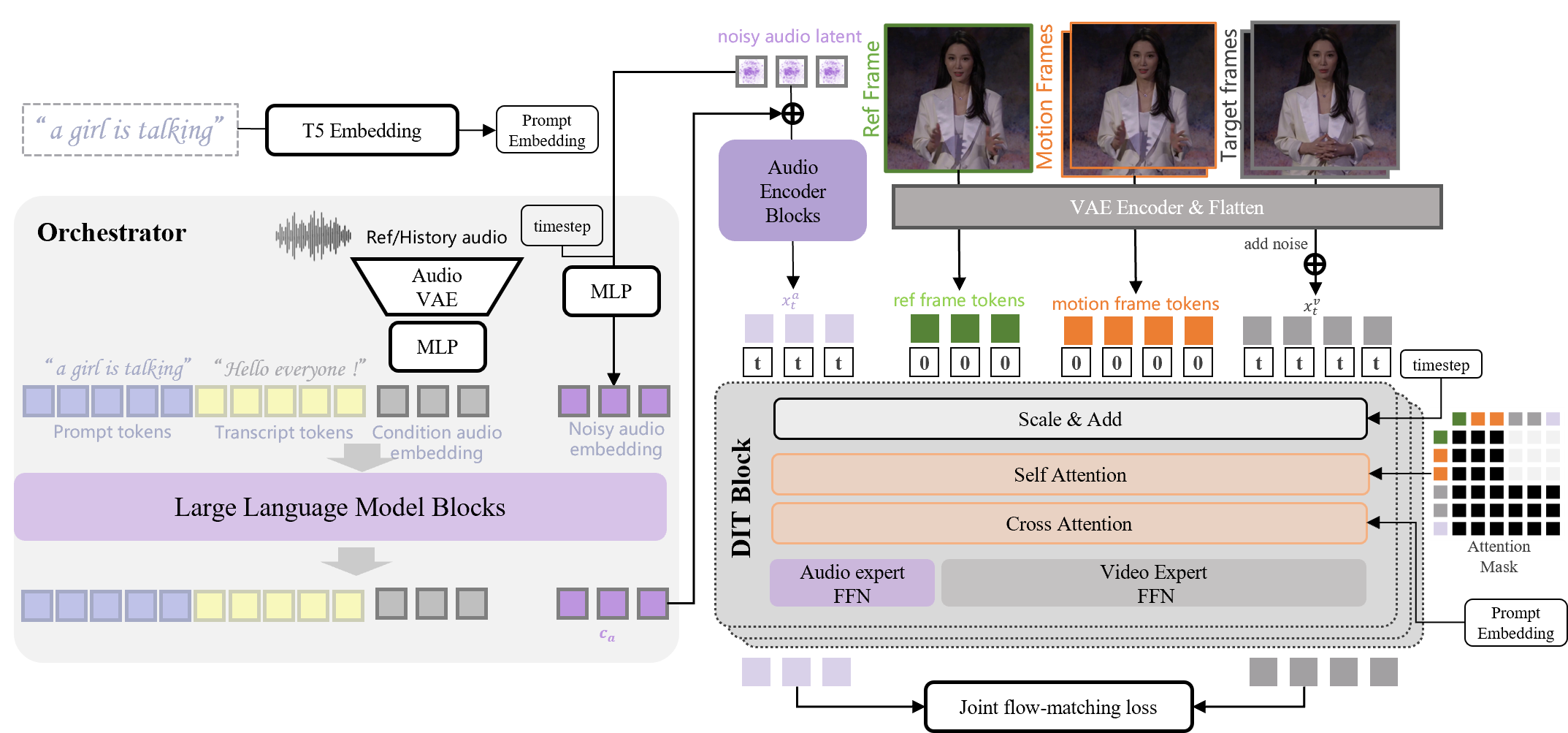}
  \Description{Block diagram: transcript and audio history feed an LLM orchestrator
    that outputs frame-aligned audio conditions to a short-window joint audio--video DiT.}
  \caption{Overall architecture. }
  \label{fig:overview}
\end{figure*}

Figure~\ref{fig:overview} illustrates the overall structure of StreamChar. The \textbf{Orchestrator} reads the prompt/transcript
and reference/history audio, producing a
frame-aligned continuous audio condition $\mathbf{c}_a$ for the active chunk. A joint audio-video
DiT then denoises $\mathbf{x}_t^{v}$ and $\mathbf{x}_t^{a}$ conditioned on prompt embeddings, $\mathbf{c}_a$, and visual conditions, including reference and motion frames. This separation keeps global transcript planning in the Orchestrator and local audio-video synthesis in the DiT.

\paragraph{Preprocessing.}
Ground-truth video is mapped to VAE latents~$\mathbf{z}_v$ with shape
$C_v\allowbreak\times T_v\allowbreak\times H'\allowbreak\times W'$;
audio is represented as audio-VAE latents~$\mathbf{z}_a \in \mathbb{R}^{T_a \times C_a}$.
Auxiliary latents include temporal motion context and the reference frame. The prompt for visual semantics
is encoded by a frozen T5 into context vectors $\mathbf{h}_{\text{T5}}$ used by the DiT.
For training, video and audio latents follow a shared \emph{latent flow} setup: each
training example defines a forward path that linearly interpolates clean latents toward
Gaussian noise in latent space. The generative objective is to learn the time-dependent
velocity field that inverts this corruption---equivalently, to denoise along the same
trajectory---so that sampling starts from noise and integrates back to plausible
$\mathbf{z}_v$ and $\mathbf{z}_a$ under prompt and Orchestrator conditioning (the DiT's regression
target is made explicit below). Concretely, we draw a time $t \in [0,1]$ and form the
intermediate states
\begin{equation}
  \mathbf{x}_t^{v} = (1-t)\,\mathbf{z}_v + t\,\boldsymbol{\epsilon}_v,
  \qquad
  \mathbf{x}_t^{a} = (1-t)\,\mathbf{z}_a + t\,\boldsymbol{\epsilon}_a,
  \label{eq:flow-latent}
\end{equation}
with $\boldsymbol{\epsilon}_v,\boldsymbol{\epsilon}_a \sim \mathcal{N}(\mathbf{0},\mathbf{I})$,
so $t{=}0$ recovers clean latents and $t{=}1$ yields pure noise. The same $t$ is used for
timestep conditioning in the Orchestrator and the DiT.

\paragraph{Orchestrator.}
An LLM pathway serves as the Orchestrator: it does not solve diffusion itself. It
reads the prompt and script, and can additionally use reference audio/text to anchor
speaker timbre, then outputs a frame-aligned audio condition
$\mathbf{c}_a$ for DiT denoising. To preserve long-form continuity, it also encodes
history from long-term generated clips.
The Orchestrator is intentionally coupled to the denoising state: it receives the
noisy audio latent $\mathbf{x}_t^a$ and the shared timestep $t$.

\paragraph{DiT.}
The DiT takes $\mathbf{x}_t^{v}$, $\mathbf{x}_t^{a}$, $t$, $\mathbf{h}_{\text{T5}}$,
the injected $\mathbf{c}_a$, and auxiliary conditional latents including motion frames,
and the reference frame.
The reference frame anchors identity and temporal consistency, while motion frames are taken from previously generated frames to improve cross-chunk coherence. We train the network $\mathbf{f}_\theta$
to regress the flow-matching velocity target
$\mathbf{v}=\boldsymbol{\epsilon}-\mathbf{z}$ for both modalities:
\begin{equation}
  \begin{split}
    \mathcal{L}_{\text{DiT}}
    &= \mathbb{E}\Big[
        \big\|\mathbf{f}_\theta^{v}(\cdot) - (\boldsymbol{\epsilon}_v - \mathbf{z}_v)\big\|_2^2 \\
        &\quad + \big\|\mathbf{f}_\theta^{a}(\cdot) - (\boldsymbol{\epsilon}_a - \mathbf{z}_a)\big\|_2^2
      \Big],
  \end{split}
  \label{eq:dit-flow-loss}
\end{equation}
where $(\cdot)$ denotes all inputs listed above.

\subsection{LLM Orchestration}
\label{sec:llmorchestration}

A causal language model serves as the \textbf{Orchestrator}. It reads the transcript and
long-term history audio features, then produces a continuous audio condition $\mathbf{c}_a$ for the DiT.

\paragraph{Continuous conditioning without an audio tokenizer.}
We do not autoregress discrete neural-codec or speech tokens to form $\mathbf{c}_a$.
The orchestrator consumes a single causal sequence of embedding vectors, packed in
fixed order:
\begin{equation}
  \mathbf{u}_{1:L}
  =
  \bigl[
    \mathbf{e}_{\mathrm{ref}},\,
    \mathbf{E}_{\mathrm{txt}},\,
    \mathbf{e}_{\mathrm{hist}},\,
    \mathbf{E}_{\mathrm{cond}}(t)
  \bigr],
  \label{eq:orchestrator-seq}
\end{equation}
where commas denote concatenation along the token axis.
Reference and history waveforms are passed through the audio VAE to obtain frame-aligned vectors. Those vectors are linearly
projected into the LLM and placed in $\mathbf{e}_{\mathrm{ref}}$ and
$\mathbf{e}_{\mathrm{hist}}$ as audio embedding blocks.
$\mathbf{E}_{\mathrm{txt}}$ is the text block: prompt and transcript tokenized by the original txt tokenizer.
$\mathbf{E}_{\mathrm{cond}}(t)$ is the conditioning tail for the current denoise step, formed by the noisy audio latent $\mathbf{x}_t^a$ and the timestep $t$. Finally the final-layer hidden states on the position of $\mathbf{E}_{\mathrm{cond}}(t)$ are collected to form $\mathbf{c}_a$.
$\mathbf{c}_a$ is learned end-to-end jointly with the DiT through the diffusion flow loss in Eq.~\eqref{eq:dit-flow-loss}.

\subsection{Joint Audio-Video Diffusion}
\label{sec:jointdiffusion}

Within each diffusion step, we patchify the reference/motion visual conditions and noisy
audio--video latents into token streams, then concatenate them as
\begin{equation}
  \mathbf{s}=[\mathbf{z}_{\mathrm{ref}},\mathbf{z}_{\mathrm{mot}},\mathbf{x}^{\,t}_{v},\mathbf{x}^{\,t}_{a}],
  \label{eq:dit-input}
\end{equation}
where $\mathbf{x}^{\,t}_{v}$ and $\mathbf{x}^{\,t}_{a}$ denote noisy video and audio tokens
at diffusion step $t$.

\paragraph{Audio alignment and joint attention.}
Before entering the main DiT blocks, audio condition features $\mathbf{c}_a$ are fused with
the noisy audio latents through a lightweight Transformer-based audio encoder, producing aligned
audio tokens. In the main DiT blocks, the denoiser applies shared self-attention over noisy video and noisy audio tokens, allowing lip motion,
prosody, and scene dynamics to interact directly at the token level. After the main DiT blocks, the audio tokens are passed through an audio decoder to obtain the final output $\mathbf{f}_\theta^{a}$.

\paragraph{Modality-Aware Mixture-of-Experts.}
To balance cross-modal interaction with modality-specific feature learning, 
our transformer blocks employ a shared attention mechanism coupled with a modality-aware Mixture-of-Experts (MoE) feed-forward network. While attention projections are shared across modalities to facilitate robust audio-visual communication, video and audio tokens are dynamically routed to distinct FFN experts within a two-expert architecture.

\paragraph{Timestep-Invariant Conditioning and Asymmetric Masking.}
For condition frames serving as reference and motion controls, we enforce timestep invariance by using clean-state embeddings ($t=0$), so these tokens represent static guidance independent of the diffusion noise schedule. To preserve this property, we introduce an asymmetric attention mask: noisy latent tokens attend to condition tokens, whereas condition tokens are masked from attending to noisy inputs. This unidirectional flow prevents stochastic noise from corrupting control signals and makes the key-value states of condition tokens invariant after the initial denoising step. Consequently, these states can be pre-computed and cached throughout the sampling trajectory, reducing redundant computation during multi-step inference.


\paragraph{Modality-aware RoPE.} 
Cross-modal attention requires precise temporal alignment despite differing latent sampling rates. In our setup, video (24\,fps, $4\times$ VAE compression) and audio (49,152\,Hz, $2048\times$ VAE compression) yield latent rates of 6\,fps and 24\,Hz, respectively, resulting in a 4:1 token density ratio. We align the modalities by scaling the RoPE base frequency for audio by $1/4$ relative to video, ensuring tokens from the same physical timestamp share identical rotational phases. For chunk-wise streaming, we maintain global temporal continuity via an offset-aware indexing scheme. Instead of resetting positions at chunk boundaries, we anchor newly generated frames at index 0 and assign negative temporal position offsets to motion-frame latents (e.g., $-K, \dots, -1$ for $K$ motion frames). This design preserves a consistent global timeline across chunks.

\begin{figure*}[t]
 \begin{center}
  \includegraphics[width=0.8\textwidth]{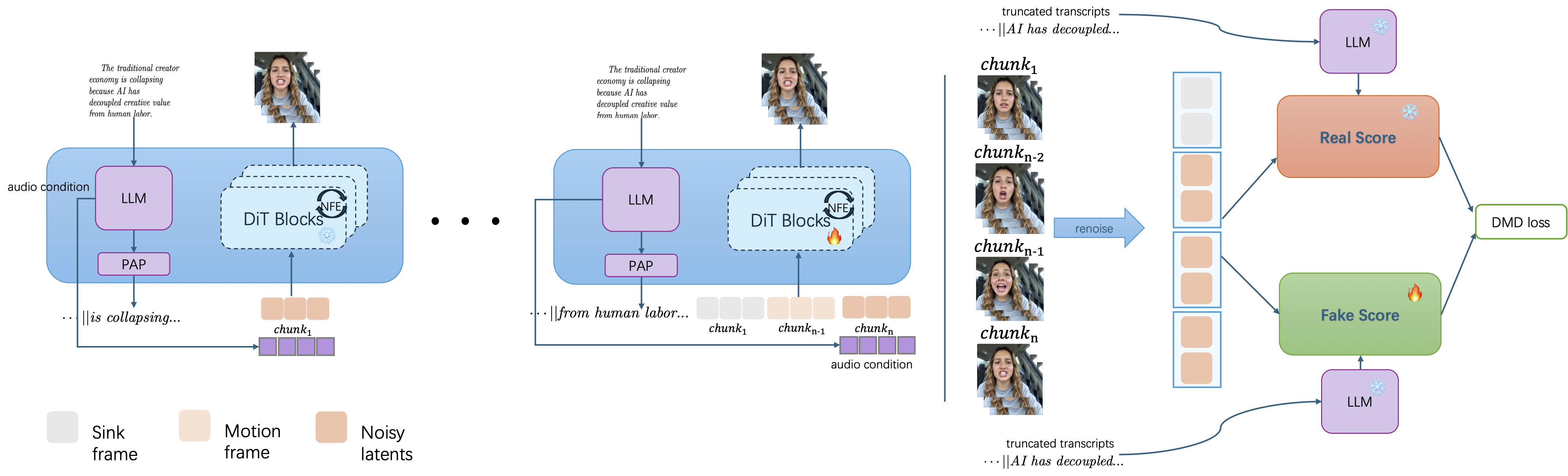}
 \end{center}
 \caption{\textbf{Online rollout distillation.} The student autoregressively 
generates $K$ chunks, with the first chunk reused as a sink memory for later 
chunks. The final three chunks are used for the DMD loss, while the 
Progress-Aware Pointer (PAP) truncates the transcript to match their audio 
progress.}
 \label{fig:distill-streaming-overview}
\end{figure*}

\begin{figure}[t]
 \begin{center}
  \includegraphics[width=0.4\textwidth]{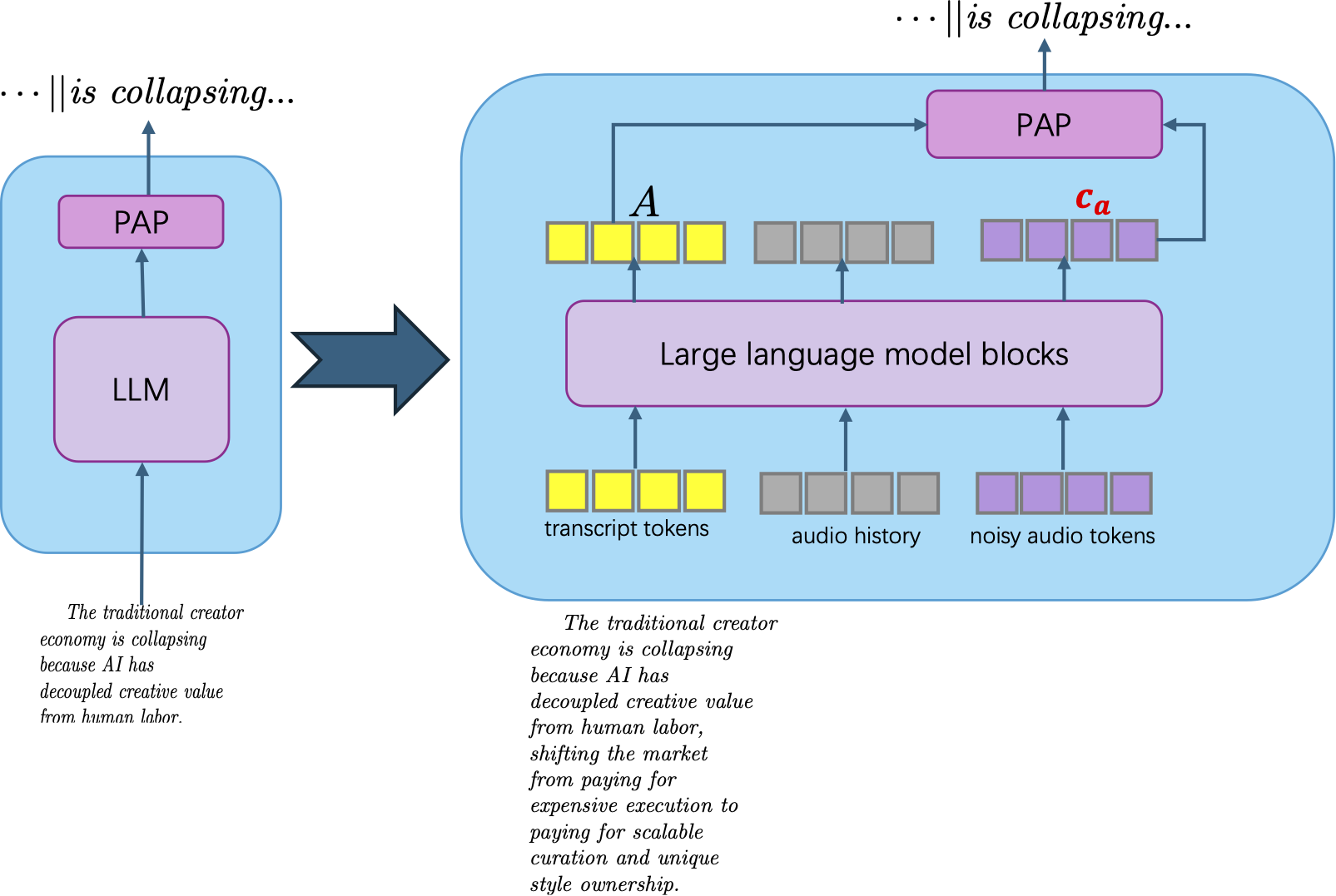}
 \end{center}
\caption{Progress-Aware Pointer (PAP). PAP predicts the spoken transcript 
endpoint from transcript states and the audio condition $c_a$; 
``\textbf{||}'' marks this endpoint for transcript truncation during rollout.}
 \label{fig:pap}
\end{figure}

\section{Streaming Inference and Distillation}
\label{sec:distill-stream}

\subsection{Bidirectional Architecture for Chunk-wise Streaming}
\label{sec:bidirectional-streaming}

Many streaming video generators adopt \emph{causal} temporal attention, where each token depends only on past frames. While this simplifies unbounded autoregression, we observe that causal training systematically \emph{degrades} generation quality because the model never accesses future context within the active clip during denoising. To preserve high-fidelity synthesis while supporting long-form streaming, we make a different design choice: we retain \emph{bidirectional} self-attention both during training and inference within each chunk.

Our DiT applies full bidirectional attention over all noisy video and audio tokens in the current window, enabling the denoiser to leverage global temporal context within the segment. This avoids the quality loss typical of purely history-conditioned causal models, which must predict each frame without seeing how subsequent frames will evolve. Long-form continuity is instead handled \emph{across} chunks through explicit conditioning on historical information rather than through architectural causality constraints.

 Concretely, during training we prepend \emph{motion} frame latents from previously generated (or ground-truth) video ahead of the noisy latents to be denoised (as defined in Eq.~\ref{eq:dit-input} and discussed in Sec.~\ref{sec:methodoverview}).
 
At inference, this same interface supports chunk-wise streaming: the student rolls forward in time by feeding motion latents from prior decoded output as $\mathbf{z}_{\mathrm{mot}}$, while bidirectional attention remains confined to the current chunk.

To bound interactive latency, we generate 9 latent video frames per chunk, which decode to 33 RGB frames under our tokenizer's temporal packing (VAE stride). At $24$\,fps, the DiT's theoretical per-chunk latency is approximately $33/24 \approx 1.38$\,s before codec and orchestration overhead---short enough for responsive streaming while still exploiting bidirectional context within the window. This design yields substantially improved visual quality compared to causal baselines, particularly in scenes requiring coherent motion planning across multiple frames.


\subsection{Two-Stage Decoupled Distillation}
\label{sec:two-stage-distill}

While bidirectional attention preserves quality within chunks, streaming inference over extended sequences introduces another challenge: \emph{error accumulation} from autoregressive generation. Naive distillation that jointly optimizes for both step reduction and long-horizon consistency leads to severe instability. We address this through a \emph{two-stage} distillation strategy that decouples step compression from rollout consistency training.

\paragraph{Stage I: Step Reduction via Distribution Matching Distillation.}
In the first stage, we distill the pretrained joint backbone into a \textbf{four-step} generator using \emph{distribution matching distillation} (DMD). This stage compresses the original 50-step sampler into a few-step generator while preserving single-chunk generation quality. On a single H100 GPU, the distilled student achieves approximately $24$\,fps generation at the clip level (including the reduced diffusion loop). The resulting four-step model serves as initialization for Stage II.

\paragraph{Stage II: Online Rollout for Autoregressive Consistency.}
Starting from the four-step initialization, we continue training with an \emph{online rollout} procedure that simulates the chunk-by-chunk generation process used at inference. During this phase, the student autoregressively generates multiple consecutive chunks, and the motion latents for later chunks are taken from the \emph{student's own} forward passes rather than solely from teacher or ground-truth crops. This exposes the optimization to the same autoregressive pipeline encountered during deployment, narrowing the train--test gap.


A key component enabling this rollout is the \textbf{progress-aware pointer (PAP)}, which determines the transcript truncation point for each chunk (Fig.~\ref{fig:pap}). PAP is integrated into the Orchestrator and trained jointly during the \emph{pretraining stage}. Given transcript hidden states $\mathbf{A}$ and audio conditions $\mathbf{c}_a$, PAP computes cross-attention to derive frame-wise soft positions $p_j$. These are refined by a learnable offset $\delta_j$ and aggregated via confidence weights $w_j$ to predict the spoken endpoint index $\hat{s}$:
\begin{equation}
  \hat{s} = \sum_j w_j (p_j + \delta_j),
\end{equation}
where $\hat{s}$ is clamped to $[0, N]$. The module is supervised using ground-truth end indices derived from ASR timestamps via smooth $\ell_1$ loss. This ensures precise alignment between the generated audio span and the transcript, allowing accurate transcript truncation for DMD loss computation in Stage II.

Another key design is the use of \emph{sink frames}: the first chunk generated by the student is persistently attended to by all subsequent chunks, providing long-range temporal memory that reduces video drifting over extended rollouts. After generating multiple chunks, we concatenate the last several chunks and feed the resulting sequence into the real-score and fake-score branches to compute the DMD loss (Fig.~\ref{fig:distill-streaming-overview}).

\section{Experiments}
\subsection{Implementation Details}
Our DiT is built upon the WAN 2.2-5B architecture~\cite{wan2025wanopenadvancedlargescale}, where we replicate the feed-forward modules to construct the audio experts. The Orchestrator is initialized from the Qwen2.5-3B architecture~\cite{qwen2025qwen25technicalreport}. Training proceeds in two phases: pretraining and distillation. 

\paragraph{Pretraining.} 
In \textbf{Pretraining Stage 1}, we pretrain the Orchestrator on the Emilia dataset~\cite{he2024emiliaextensivemultilingualdiverse} for 80k steps, using a batch size of 640 and a learning rate of $6 \times 10^{-5}$. In \textbf{Pretraining Stage 2}, we curate an audio-video dataset by combining SpeakerVid-5M~\cite{zhang2025speakervid5mlargescalehighqualitydataset}, TalkVid~\cite{chen2025talkvidlargescalediversifieddataset}, and OpenHumanVid~\cite{li2025openhumanvidlargescalehighqualitydataset}. We then jointly train the Orchestrator and DiT for 100k steps with a batch size of 128 and a learning rate of $1 \times 10^{-5}$. During generation, the model produces 33 frames per chunk at 24 fps. To ensure seamless chunk-by-chunk synthesis, we align the number of motion frames with the chunk size (33 frames).
The maximum duration of the historical audio input to the Orchestrator is set to 15 seconds. Since our training data contains no videos/transcripts longer than 20 seconds, we truncate the oldest audio segments and their corresponding transcripts during inference to ensure consistency between training and inference.

\paragraph{Distillation.}
To enable real-time streaming, we employ a two-stage distillation strategy distinct from pretraining. In \textbf{Distillation Stage I} (step compression), we train for 600 steps to compress the sampler into a 4-step generator. In \textbf{Distillation Stage II} (online rollout consistency), we train for 400 steps to refine autoregressive stability. For both distillation stages, the student network uses a learning rate of $2 \times 10^{-6}$, while the fake score network uses a learning rate of $4 \times 10^{-7}$.

\paragraph{Inference Efficiency.}
On a single H100 GPU, generating a 33-frame chunk ($512\times512$, 4 steps) in bfloat16 takes \textbf{0.96\,s} (LLM + DiT). The pipeline adds VAE decoding (\textasciitilde0.30\,s), preprocessing (\textasciitilde0.05\,s), and stream writing (0.025\,s). Crucially, motion frame latents are directly reused from the preceding chunk, bypassing VAE encoding. Furthermore, the next chunk's preprocessing initiates immediately after generation, overlapping with VAE decoding to streamline chunk transitions. The total per-chunk latency (\textasciitilde1.34\,s) remains within the playback budget ($33/24 \approx 1.38$\,s).

\subsection{Compared Methods}
To the best of our knowledge, open-source methods capable of long-horizon real-time streaming joint audio-video generation remain highly limited. While concurrent work such as Hallo-Live~\cite{li2026halloliverealtimestreamingjoint} explores real-time joint streaming, it is primarily confined to short durations. We therefore adopt a two-tier comparison strategy and explicitly separate task comparability from system capability. 
First, we evaluate against recent open-source real-time audio-driven video generation methods: SoulX-FlashTalk~\cite{shen2026soulxflashtalkrealtimeinfinitestreaming}, SoulX-FlashHead~\cite{yu2026soulxflashheadoracleguidedgenerationinfinite}, and LiveAvatar~\cite{huang2026liveavatarstreamingrealtime}. These methods do not synthesize speech from text, but they are the closest references for low-latency streaming video generation. 
Second, we include state-of-the-art open-source audio-video generation models, namely LTX-2~\cite{hacohen2026ltx2efficientjointaudiovisual}, OVI~\cite{low2025ovitwinbackbonecrossmodal}, and MagiHuman~\cite{davinci-magihuman-2026}. These models operate in a non-streaming manner, but provide strong references for perceptual quality, temporal coherence, and cross-modal alignment. 
This comparative framework evaluates both the streaming efficiency and the multimodal generation fidelity of StreamChar, while making explicit which baselines solve the full text-to-audio-video problem and which evaluate only the video-generation subproblem.

Since our model is a unified audio-video generator that produces both speech and visual motion from text, we use its generated audio as input to the streaming audio-driven baselines. This protocol keeps the driving audio identical when evaluating video synthesis, although these baselines do not solve the same text-to-audio-video task. We report results on two protocols derived from the EMTD dataset~\cite{meng2026echomimicv2strikingsimplifiedsemibody}: a standard set of 150 clips generating 10s audio-video pairs from original transcripts and first frames, and a long-horizon set of 50 clips paired with randomly sampled transcripts ($>$300 words) to produce 5-minute continuous streams. Across both settings, our synthesized audio serves as the sole driving signal for the streaming methods.

\subsection{Quantitative Comparison}
\paragraph{Evaluation metrics.} 
Audio--visual synchronization is measured via Sync-C and Sync-D~\cite{chung2016out}. 
Perceptual fidelity is quantified by FID and FVD~\cite{heusel2017gans,unterthiner2019towards}. 
Human-centric quality uses VBench-2.0's \emph{Human Anatomy} and \emph{Human Identity} dimensions~\cite{zheng2025vbench20}. 
Speech intelligibility and transcript alignment are captured by WER (Word Error Rate). 
Audio-driven baselines omit WER as they do not synthesize speech. 
For long-horizon streaming stability, we additionally report VBench's \emph{Dynamic} score (motion diversity) and a \emph{Quality Drift} metric. Following rolling-forcing~\cite{liu2025rolling}, every 30 seconds we compute the absolute quality difference between that segment's final 5 seconds and the video's initial 5 seconds, and report the maximum difference over the full video as drift.
In Table~\ref{tab:quant-main}, \textbf{bold} and \underline{underlined} values denote the best and second-best results per metric, respectively. 
Note that "Ours (base model)" refers to the model after pretraining, while "Ours (after distill)" denotes the 4-step student after DMD and online rollout distillation.

\begin{table*}[t]
  \centering
  \small
  \caption{Quantitative comparison on the EMTD benchmark. \textbf{Bold} and \underline{underlined} values denote the best and second-best results per metric. Ablation studies are shown for reference but do not compete for rankings.}
  \label{tab:quant-main}
  \begin{tabular}{@{}lccccccc@{}}
    \toprule
    Method & Sync-C$\uparrow$ & Sync-D$\downarrow$  & FID $\downarrow$ & FVD $\downarrow$ & H.\,Anat.\,$\uparrow$ & H.\,Id.\,$\uparrow$ & WER (\%) $\downarrow$ \\
    \midrule
    OVI~\cite{low2025ovitwinbackbonecrossmodal}
      & 7.183 & 8.692 & 21.092 & 287.88 & 0.929 & 0.892 & 10.458 \\
    LTX-2~\cite{hacohen2026ltx2efficientjointaudiovisual}
      & 7.892 & 7.979 & 23.019 & 275.68 & 0.912 & 0.921 & 4.549 \\
    MagiHuman~\cite{davinci-magihuman-2026}
      & \underline{8.754} & \textbf{7.156} & 18.122 & \textbf{235.97} & 0.913 & \textbf{0.981} & 7.717 \\
    \midrule
    Ours (base model)
      & 7.427 & 8.309 & \underline{17.987} & \underline{248.16} & \underline{0.939} & 0.941 & \textbf{3.539} \\
    Ours (after distill)
      & 8.126 & 8.497 & 18.963 & 289.091 & 0.941 & 0.924 & \underline{3.649} \\
    \midrule
    SoulX-FlashTalk~\cite{shen2026soulxflashtalkrealtimeinfinitestreaming}
      & \textbf{9.067} & \underline{7.461} & \textbf{14.982} & 278.45 & 0.923 & \underline{0.973} & -- \\
    SoulX-FlashHead~\cite{yu2026soulxflashheadoracleguidedgenerationinfinite}
      & 7.866 & 8.022 & 18.907 & 314.55 & \textbf{0.957} & 0.970 & -- \\
    LiveAvatar~\cite{huang2026liveavatarstreamingrealtime}
      & 7.204 & 8.556 & 20.392 & 394.27 & 0.924 & 0.979 & -- \\
    \midrule
    Ours (stage-2 one-chunk) 
      & 5.596 & 9.280 & 19.453 & 285.45 & 0.937 & 0.923 & 35.436 \\
    Ours (distill stage-2 only)   
      & 6.788 & 9.455 & 17.446 & 265.40 & 0.948 & 0.942 & 5.756 \\
    \bottomrule
  \end{tabular}
\end{table*}

\begin{table}[t]
  \centering
  \small
  \caption{Long-horizon streaming evaluation.}
  \label{tab:long-term}
  \begin{tabular}{@{}lcccc@{}}
    \toprule
    Method & Sync-C & Sync-D & Dynamic$\uparrow$ & Drift $\downarrow$ \\
    \midrule
    Ours w/o sink chunk & 8.052 & 8.180 & 1.0 & 0.0304 \\
    \textbf{Ours} & 8.185 & 8.388 & 1.0 & 0.0067 \\
    FlashTalk & 9.593 & 7.249 & 0.75 & 0.0055 \\
    FlashHead & 7.419 & 8.685 & 0.05 & 0.0088 \\
    LiveAvatar & 7.983 & 8.005 & 1.0 & 0.0130 \\
    \bottomrule
  \end{tabular}
\end{table}

\paragraph{Results and Analysis.}
Our method demonstrates an advantage in speech intelligibility and transcript alignment, as evidenced by the WER metric. The base model (3.54\%) outperforms the evaluated joint audio-video generators, while the chunk-wise distilled variant maintains near-identical performance (3.65\%). This suggests that the LLM orchestrator preserves fine-grained phonetic alignment during continuous, multi-chunk streaming generation. By offloading global script understanding and acoustic intent planning to the LLM, the DiT can focus on short-window denoising while receiving chunk-level acoustic guidance.

In terms of visual fidelity, our method remains competitive despite employing a more compact backbone. Built upon a \textbf{5B-parameter} video foundation model, our approach achieves perceptual quality comparable to recent generators that typically scale beyond \textbf{14B parameters} (e.g., LTX-2, MagiHuman, SoulX-FlashTalk, LiveAvatar). Among joint audio-video baselines, we obtain a leading \textit{Human Anatomy} score and the lowest FID (17.99). The lower FID of specialized audio-driven pipelines (e.g., SoulX-FlashTalk) may partly reflect their more constrained motion scope, where dynamics are often concentrated around facial and hand regions while preserving spatial consistency with the reference frame. Our joint model simultaneously generates speech and upper-body motion, making the comparison more demanding. Regarding audio-visual synchronization, our Sync-C/D scores are competitive but do not dominate the specialized audio-driven baselines, which directly condition on waveforms. Following aggressive distillation (50 $\rightarrow$ 4 steps), we observe a mild increase in FVD, reflecting the expected quality-efficiency trade-off under step reduction. Importantly, synchronization and speech intelligibility (WER) remain stable, and anatomical quality is preserved. Overall, the quantitative results support a practical balance across visual realism, cross-modal alignment, transcript fidelity, and streaming efficiency.
For long-horizon streaming stability, we refer to Table~\ref{tab:long-term}. Our method achieves a negligible Quality Drift of $0.0067$, demonstrating robust suppression of error accumulation over extended sequences. Crucially, \textbf{Ours} maintains the maximum VBench \textit{Dynamic} score ($1.0$), indicating that this stability is achieved without sacrificing motion diversity or anchoring to the reference frame. LiveAvatar, despite achieving a high Dynamic score, suffers from visible oscillatory artifacts. 
While StreamChar does not dominate every individual metric, it offers the strongest overall trade-off among joint generation ability, streaming latency, transcript fidelity, and long-horizon stability.

\subsection{Qualitative Comparison}

\begin{figure*}[t]
  \centering
  \includegraphics[width=0.88\textwidth]{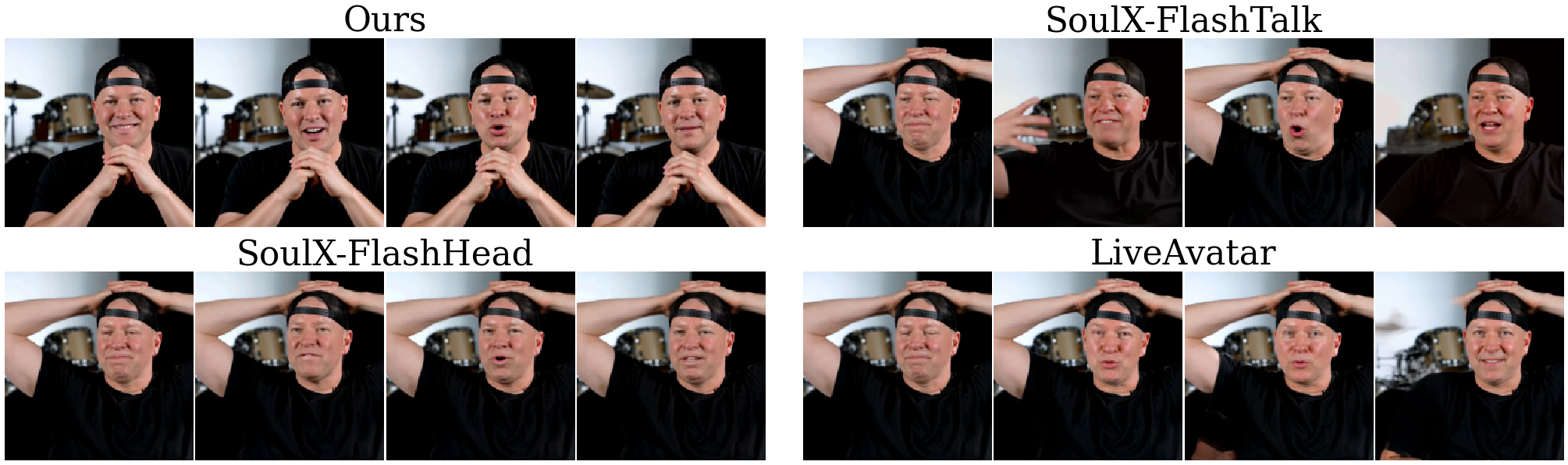}\\[-0.15ex]
  \includegraphics[width=0.88\textwidth]{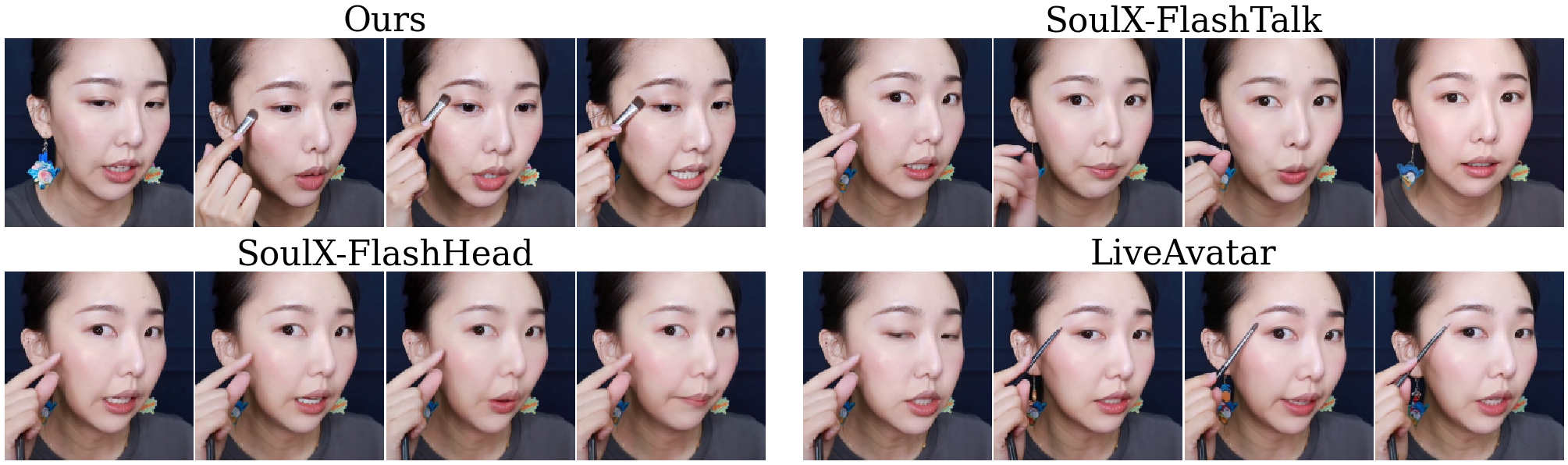}\\[-0.15ex]
  \includegraphics[width=0.88\textwidth]{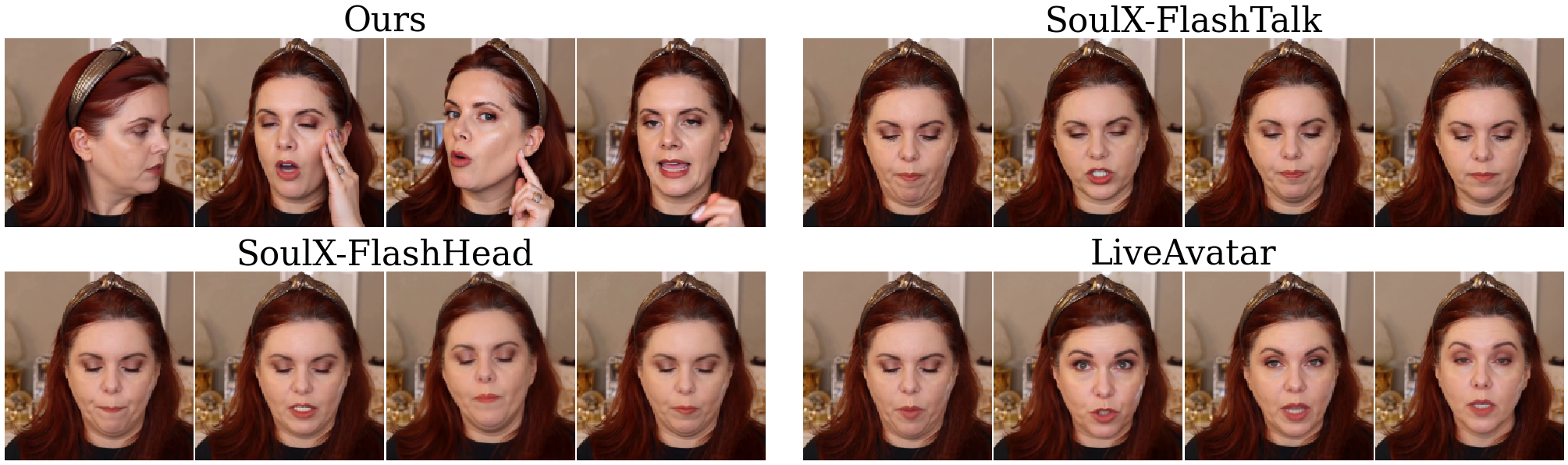}

  \caption{Qualitative comparison with SoulX-FlashTalk, SoulX-FlashHead, and LiveAvatar.}
  \label{fig:qual-comp}
\end{figure*}

Figure~\ref{fig:long-term} illustrates the generation behavior of our method over minute-scale sequences, where it maintains stable temporal continuity across extended rollouts. Together with Figure~\ref{fig:qual-comp}, both visualizations suggest a common tendency in recent streaming baselines: generated frames tend to remain closely anchored to the initial reference image. As a result, motion is often confined to localized facial expressions or hand gestures, while torso posture typically remains static or exhibits minimal variation. In comparison, our method exhibits comparatively less reliance on reference anchoring, yielding more varied upper-body poses and natural hand--object interactions, with motion that frequently extends beyond the immediate reference neighborhood. 

\begin{figure*}[t]
  \centering
  \includegraphics[width=0.95\textwidth]{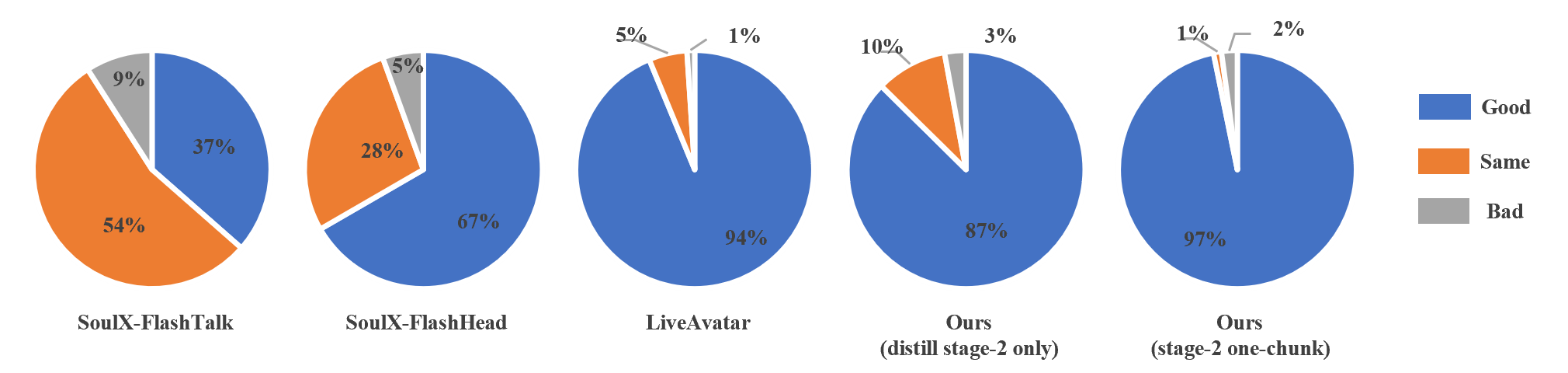}
  \caption{User study under the GSB protocol.}
  \label{fig:user-study}
\end{figure*}

\paragraph{User study.}
We further conduct a GSB (good-same-bad) user study for the streaming setting, where our full system corresponds to \texttt{Ours (after distill)}. We recruit 24 participants, each presented with 50 randomly sampled cases. In each case, two randomly selected results are shown and participants judge their relative quality in terms of motion naturalness, lip-sync accuracy, and motion richness. As shown in Figure~\ref{fig:user-study}, our method achieves more favorable preferences than competing streaming baselines and ablated variants, suggesting a perceptual advantage of the proposed design.

\subsection{Ablation Study.}
Table~\ref{tab:quant-main} further validates key design choices.

\begin{figure*}[t]
  \centering
  
  \textbf{Ours}\\[0.0ex]
  \includegraphics[width=\textwidth]{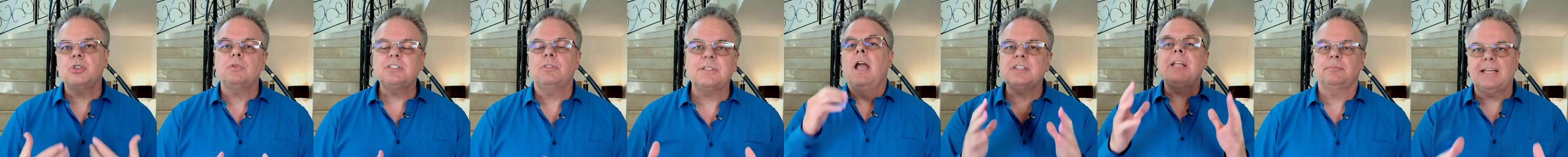}\\[-0.12ex]
  
  \textbf{Ours w/o sink chunk}\\[-0.0ex]
  \includegraphics[width=\textwidth]{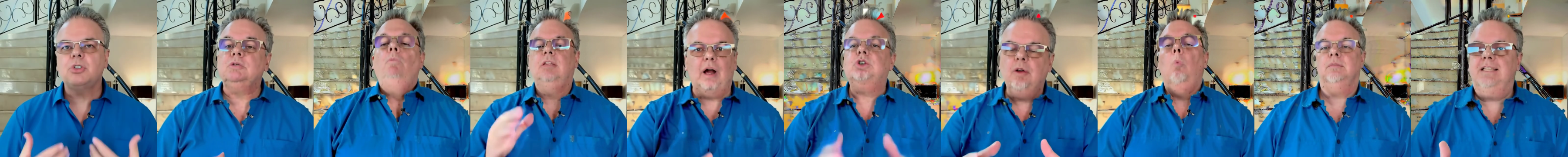}\\[-0.12ex]

  \textbf{SoulX-FlashTalk}\\[0.0ex]
  \includegraphics[width=\textwidth]{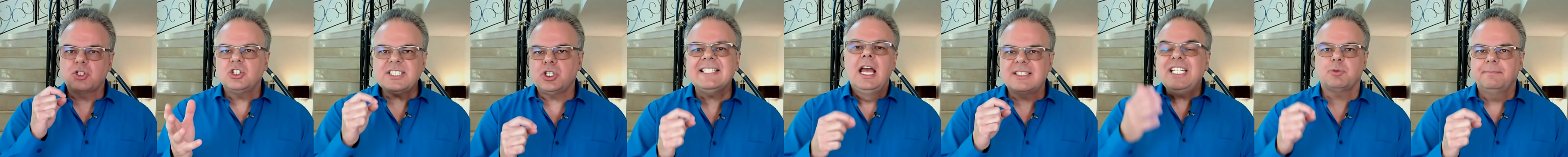}\\[-0.12ex]

    \textbf{SoulX-FlashHead}\\[0.0ex]
  \includegraphics[width=\textwidth]{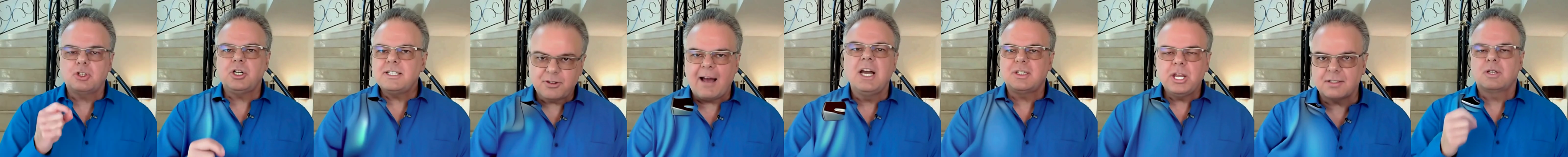}\\[-0.12ex]

    \textbf{LiveAvatar}\\[0.0ex]
  \includegraphics[width=\textwidth]{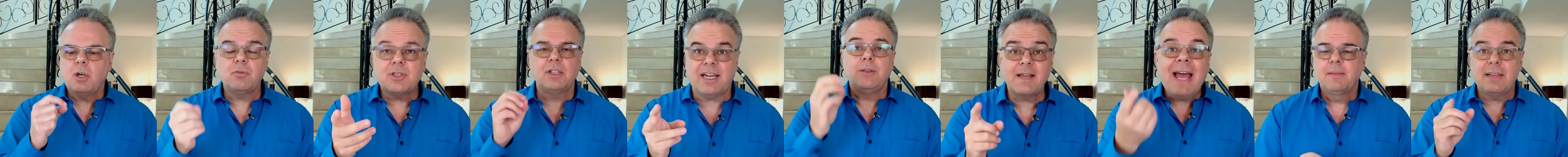}\\[-0.12ex]

  \makebox[\textwidth]{%
    \begin{tikzpicture}[x=\textwidth/300, y=1cm]
      \draw[thick] (0,0) -- (300,0);
      \draw[thick] (0,0) -- (0,-0.1) node[below, font=\small] {0s};
      \draw[thick] (100,0) -- (100,-0.1) node[below, font=\small] {100s};
      \draw[thick] (200,0) -- (200,-0.1) node[below, font=\small] {200s};
      \draw[thick] (300,0) -- (300,-0.1) node[below, font=\small] {300s};
    \end{tikzpicture}%
  }\\[1ex] 
  \vspace*{-4mm} 
  \caption{Long-horizon qualitative comparison.}
  \label{fig:long-term}
\end{figure*}

\begin{figure*}[t]
  \centering
  
  \textbf{with sink chunk}\\[0.5ex]
  \includegraphics[width=\textwidth]{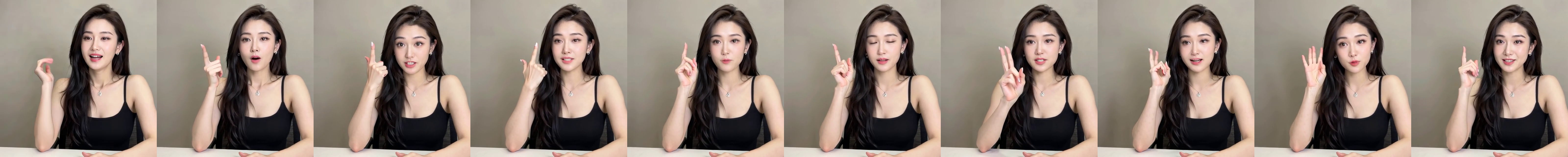}\\[-0.12ex]
  \includegraphics[width=\textwidth]{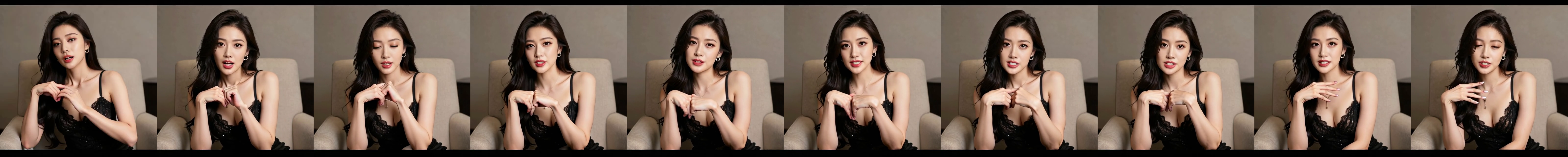}\\[1ex]
  
  \textbf{w/o sink chunk}\\[0.5ex]
  \includegraphics[width=\textwidth]{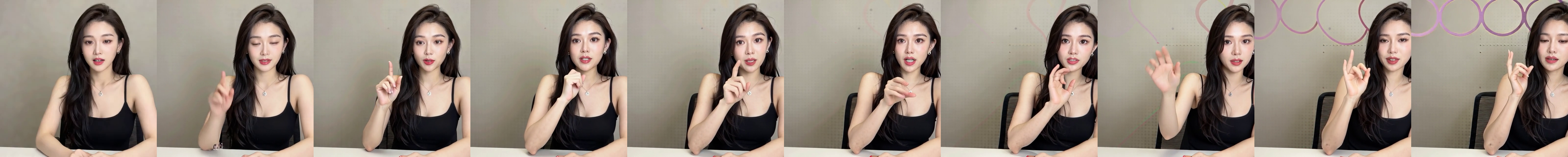}\\[-0.12ex]
  \includegraphics[width=\textwidth]{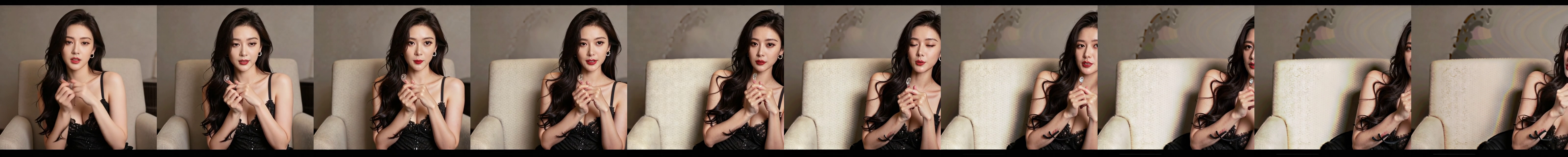}

  \makebox[\textwidth]{%
    \begin{tikzpicture}[x=\textwidth/200, y=1cm] 
      \draw[thick] (0,0) -- (200,0);
      \draw[thick] (0,0) -- (0,-0.1) node[below, font=\small] {0s};
      \draw[thick] (50,0) -- (50,-0.1) node[below, font=\small] {50s};
      \draw[thick] (100,0) -- (100,-0.1) node[below, font=\small] {100s};
      \draw[thick] (200,0) -- (200,-0.1) node[below, font=\small] {200s};
    \end{tikzpicture}%
  }\\[1ex] 
  \vspace*{-4mm} 
  \caption{Sink-chunk conditioning reduces long-horizon drift and repetitive spatial behavior.}
  \label{fig:sink-chunk}
\end{figure*}

\textbf{Sink Chunk for Error Accumulation and Mode Collapse.}
Table~\ref{tab:long-term} shows that removing the sink chunk increases Quality Drift from $0.0067$ to $0.0304$. As visualized in Figure~\ref{fig:long-term}, this metric reflects severe error accumulation across chunks, manifesting as noticeable color shifts and appearance degradation over time. Figure~\ref{fig:sink-chunk} further reveals that without sink conditioning, the distilled student tends to exhibit stereotyped spatial behaviors and persistent spatial offsets, suggesting a collapse toward low-diversity motion patterns. The sink mechanism mitigates these issues by providing a stable long-range reference, effectively suppressing quality drift and mode collapse while preserving motion diversity (Dynamic score $1.0$).

\textbf{Single-chunk vs. Multi-chunk in Stage~II:} We compare our online rollout over multiple consecutive chunks against training on isolated chunk only (\texttt{Ours (stage-2 one-chunk)}). Training on isolated chunk degrades the WER to 35.4\% and reduces Sync-C/D scores. This occurs because shorter sequences lack sufficient cross-chunk acoustic context for correct transcript alignment. Concatenating multiple chunks allows the distillation objective to capture consistent prosody and long-range phonetic transitions.

\textbf{Two-stage vs.\ single-stage distillation.} We ablate the pipeline by skipping Stage~I (DMD step compression) and training Stage~II directly from the 50-step teacher (\texttt{Ours (distill stage-2 only)}). While several short-clip metrics remain competitive, qualitative results reveal issues such as motion suppression and reference-frame anchoring. Directly combining step reduction with autoregressive rollout training places competing pressure on the student: it must learn a low-step sampler and recover from its own rollout errors at the same time. In contrast, decoupling first stabilizes the few-step mapping (Stage~I) and then refines cross-chunk consistency (Stage~II), thereby preserving both efficiency and visual dynamics.

\section{Limitations}

StreamChar is designed for long-horizon, text-driven character generation, but several limitations remain. First, audio-driven streaming baselines are evaluated using our generated audio as the common driver, so they are controlled video references rather than full text-to-audio-video competitors. Second, real-time performance is measured on a single H100 GPU with a 33-frame chunk budget; lower-end deployment may require additional optimization. 



\section{Conclusion}

We have presented StreamChar, a decoupled LLM--DiT framework for long-horizon streaming audio--video generation. The LLM orchestrator handles transcript-level planning, while the DiT performs short-window joint denoising with motion conditioning. A two-stage distillation strategy, together with PAP and sink-frame memory, enables real-time chunk-wise generation with rollout stability. Experiments show that StreamChar runs in real time on a single GPU and offers a competitive balance of transcript fidelity, audio-visual synchronization, visual quality, and streaming stability.

%
%
%
%


\bibliographystyle{ACM-Reference-Format}
\bibliography{ravgen-references}

@inproceedings{peebles2023scalable,
  author    = {Peebles, William and Xie, Saining},
  title     = {Scalable Diffusion Models with Transformers},
  booktitle = {Proceedings of the IEEE/CVF International Conference on Computer Vision (ICCV)},
  year      = {2023},
  pages     = {4195--4205}
}

@inproceedings{rombach2022high,
  author    = {Rombach, Robin and Blattmann, Andreas and Lorenz, Dominik and Esser, Patrick and Ommer, Bj{\"o}rn},
  title     = {High-Resolution Image Synthesis with Latent Diffusion Models},
  booktitle = {Proceedings of the IEEE/CVF Conference on Computer Vision and Pattern Recognition (CVPR)},
  year      = {2022},
  pages     = {10684--10695}
}

@inproceedings{ho2020denoising,
  author    = {Ho, Jonathan and Jain, Ajay and Abbeel, Pieter},
  title     = {Denoising Diffusion Probabilistic Models},
  booktitle = {Advances in Neural Information Processing Systems (NeurIPS)},
  volume    = {33},
  year      = {2020}
}

@inproceedings{vaswani2017attention,
  author    = {Vaswani, Ashish and Shazeer, Noam and Parmar, Niki and Uszkoreit, Jakob and Jones, Llion and Gomez, Aidan N. and Kaiser, {\L}ukasz and Polosukhin, Illia},
  title     = {Attention Is All You Need},
  booktitle = {Advances in Neural Information Processing Systems (NeurIPS)},
  volume    = {30},
  year      = {2017}
}

@misc{wan2025wanopenadvancedlargescale,
      title={Wan: Open and Advanced Large-Scale Video Generative Models}, 
      author={Team Wan and Ang Wang and Baole Ai and Bin Wen and Chaojie Mao and Chen-Wei Xie and Di Chen and Feiwu Yu and Haiming Zhao and Jianxiao Yang and Jianyuan Zeng and Jiayu Wang and Jingfeng Zhang and Jingren Zhou and Jinkai Wang and Jixuan Chen and Kai Zhu and Kang Zhao and Keyu Yan and Lianghua Huang and Mengyang Feng and Ningyi Zhang and Pandeng Li and Pingyu Wu and Ruihang Chu and Ruili Feng and Shiwei Zhang and Siyang Sun and Tao Fang and Tianxing Wang and Tianyi Gui and Tingyu Weng and Tong Shen and Wei Lin and Wei Wang and Wei Wang and Wenmeng Zhou and Wente Wang and Wenting Shen and Wenyuan Yu and Xianzhong Shi and Xiaoming Huang and Xin Xu and Yan Kou and Yangyu Lv and Yifei Li and Yijing Liu and Yiming Wang and Yingya Zhang and Yitong Huang and Yong Li and You Wu and Yu Liu and Yulin Pan and Yun Zheng and Yuntao Hong and Yupeng Shi and Yutong Feng and Zeyinzi Jiang and Zhen Han and Zhi-Fan Wu and Ziyu Liu},
      year={2025},
      eprint={2503.20314},
      archivePrefix={arXiv},
      primaryClass={cs.CV},
      url={https://arxiv.org/abs/2503.20314}, 
}

@misc{qwen2025qwen25technicalreport,
      title={Qwen2.5 Technical Report}, 
      author={Qwen and : and An Yang and Baosong Yang and Beichen Zhang and Binyuan Hui and Bo Zheng and Bowen Yu and Chengyuan Li and Dayiheng Liu and Fei Huang and Haoran Wei and Huan Lin and Jian Yang and Jianhong Tu and Jianwei Zhang and Jianxin Yang and Jiaxi Yang and Jingren Zhou and Junyang Lin and Kai Dang and Keming Lu and Keqin Bao and Kexin Yang and Le Yu and Mei Li and Mingfeng Xue and Pei Zhang and Qin Zhu and Rui Men and Runji Lin and Tianhao Li and Tianyi Tang and Tingyu Xia and Xingzhang Ren and Xuancheng Ren and Yang Fan and Yang Su and Yichang Zhang and Yu Wan and Yuqiong Liu and Zeyu Cui and Zhenru Zhang and Zihan Qiu},
      year={2025},
      eprint={2412.15115},
      archivePrefix={arXiv},
      primaryClass={cs.CL},
      url={https://arxiv.org/abs/2412.15115}, 
}

@misc{he2024emiliaextensivemultilingualdiverse,
      title={Emilia: An Extensive, Multilingual, and Diverse Speech Dataset for Large-Scale Speech Generation}, 
      author={Haorui He and Zengqiang Shang and Chaoren Wang and Xuyuan Li and Yicheng Gu and Hua Hua and Liwei Liu and Chen Yang and Jiaqi Li and Peiyang Shi and Yuancheng Wang and Kai Chen and Pengyuan Zhang and Zhizheng Wu},
      year={2024},
      eprint={2407.05361},
      archivePrefix={arXiv},
      primaryClass={eess.AS},
      url={https://arxiv.org/abs/2407.05361}, 
}

@misc{zhang2025speakervid5mlargescalehighqualitydataset,
      title={SpeakerVid-5M: A Large-Scale High-Quality Dataset for Audio-Visual Dyadic Interactive Human Generation}, 
      author={Youliang Zhang and Zhaoyang Li and Duomin Wang and Jiahe Zhang and Deyu Zhou and Zixin Yin and Xili Dai and Gang Yu and Xiu Li},
      year={2025},
      eprint={2507.09862},
      archivePrefix={arXiv},
      primaryClass={cs.CV},
      url={https://arxiv.org/abs/2507.09862}, 
}

@misc{chen2025talkvidlargescalediversifieddataset,
      title={TalkVid: A Large-Scale Diversified Dataset for Audio-Driven Talking Head Synthesis}, 
      author={Shunian Chen and Hejin Huang and Yexin Liu and Zihan Ye and Pengcheng Chen and Chenghao Zhu and Michael Guan and Rongsheng Wang and Junying Chen and Guanbin Li and Ser-Nam Lim and Harry Yang and Benyou Wang},
      year={2025},
      eprint={2508.13618},
      archivePrefix={arXiv},
      primaryClass={cs.CV},
      url={https://arxiv.org/abs/2508.13618}, 
}

@misc{li2025openhumanvidlargescalehighqualitydataset,
      title={OpenHumanVid: A Large-Scale High-Quality Dataset for Enhancing Human-Centric Video Generation}, 
      author={Hui Li and Mingwang Xu and Yun Zhan and Shan Mu and Jiaye Li and Kaihui Cheng and Yuxuan Chen and Tan Chen and Mao Ye and Jingdong Wang and Siyu Zhu},
      year={2025},
      eprint={2412.00115},
      archivePrefix={arXiv},
      primaryClass={cs.CV},
      url={https://arxiv.org/abs/2412.00115}, 
}

@misc{shen2026soulxflashtalkrealtimeinfinitestreaming,
      title={SoulX-FlashTalk: Real-Time Infinite Streaming of Audio-Driven Avatars via Self-Correcting Bidirectional Distillation}, 
      author={Le Shen and Qian Qiao and Tan Yu and Ke Zhou and Tianhang Yu and Yu Zhan and Zhenjie Wang and Ming Tao and Shunshun Yin and Siyuan Liu},
      year={2026},
      eprint={2512.23379},
      archivePrefix={arXiv},
      primaryClass={cs.CV},
      url={https://arxiv.org/abs/2512.23379}, 
}

@misc{yu2026soulxflashheadoracleguidedgenerationinfinite,
      title={SoulX-FlashHead: Oracle-guided Generation of Infinite Real-time Streaming Talking Heads}, 
      author={Tan Yu and Qian Qiao and Le Shen and Ke Zhou and Jincheng Hu and Dian Sheng and Bo Hu and Haoming Qin and Jun Gao and Changhai Zhou and Shunshun Yin and Siyuan Liu},
      year={2026},
      eprint={2602.07449},
      archivePrefix={arXiv},
      primaryClass={cs.CV},
      url={https://arxiv.org/abs/2602.07449}, 
}

@misc{huang2026liveavatarstreamingrealtime,
      title={Live Avatar: Streaming Real-time Audio-Driven Avatar Generation with Infinite Length}, 
      author={Yubo Huang and Hailong Guo and Fangtai Wu and Shifeng Zhang and Shijie Huang and Qijun Gan and Lin Liu and Sirui Zhao and Enhong Chen and Jiaming Liu and Steven Hoi},
      year={2026},
      eprint={2512.04677},
      archivePrefix={arXiv},
      primaryClass={cs.CV},
      url={https://arxiv.org/abs/2512.04677}, 
}

@misc{meng2026echomimicv2strikingsimplifiedsemibody,
      title={EchoMimicV2: Towards Striking, Simplified, and Semi-Body Human Animation}, 
      author={Rang Meng and Xingyu Zhang and Yuming Li and Chenguang Ma},
      year={2026},
      eprint={2411.10061},
      archivePrefix={arXiv},
      primaryClass={cs.GR},
      url={https://arxiv.org/abs/2411.10061}, 
}

@misc{low2025ovitwinbackbonecrossmodal,
      title={Ovi: Twin Backbone Cross-Modal Fusion for Audio-Video Generation}, 
      author={Chetwin Low and Weimin Wang and Calder Katyal},
      year={2025},
      eprint={2510.01284},
      archivePrefix={arXiv},
      primaryClass={cs.MM},
      url={https://arxiv.org/abs/2510.01284}, 
}

@article{davinci-magihuman-2026,
  title={Speed by Simplicity: A Single-Stream Architecture for Fast Audio-Video Generative Foundation Model},
  author={SII-GAIR and Sand. ai and Chern, Ethan and Teng, Hansi and Sun, Hanwen and Wang, Hao and Pan, Hong and Jia, Hongyu and Su, Jiadi and Li, Jin and Yu, Junjie and Liu, Lijie and Li, Lingzhi and Ye, Lyumanshan and Hu, Min and Wang, Qiangang and Qi, Quanwei and Chern, Steffi and Bu, Tao and Wang, Taoran and Xu, Teren and Zhang, Tianning and Mi, Tiantian and Xu, Weixian and Zhang, Wenqiang and Zhang, Wentai and Yi, Xianping and Cai, Xiaojie and Kang, Xiaoyang and Ma, Yan and Liu, Yixiu and Zhang, Yunbo and Huang, Yunpeng and Lin, Yutong and Tao, Zewei and Liu, Zhaoliang and Zhang, Zheng and Cen, Zhiyao and Yu, Zhixuan and Wang, Zhongshu and Hu, Zhulin and Zhou, Zijin and Guo, Zinan and Cao, Yue and Liu, Pengfei},
  journal={arXiv preprint arXiv:2603.21986},
  year={2026}
}

@misc{hacohen2026ltx2efficientjointaudiovisual,
      title={LTX-2: Efficient Joint Audio-Visual Foundation Model}, 
      author={Yoav HaCohen and Benny Brazowski and Nisan Chiprut and Yaki Bitterman and Andrew Kvochko and Avishai Berkowitz and Daniel Shalem and Daphna Lifschitz and Dudu Moshe and Eitan Porat and Eitan Richardson and Guy Shiran and Itay Chachy and Jonathan Chetboun and Michael Finkelson and Michael Kupchick and Nir Zabari and Nitzan Guetta and Noa Kotler and Ofir Bibi and Ori Gordon and Poriya Panet and Roi Benita and Shahar Armon and Victor Kulikov and Yaron Inger and Yonatan Shiftan and Zeev Melumian and Zeev Farbman},
      year={2026},
      eprint={2601.03233},
      archivePrefix={arXiv},
      primaryClass={cs.CV},
      url={https://arxiv.org/abs/2601.03233}, 
}

@inproceedings{chung2016out,
  author    = {Chung, Joon Son and Zisserman, Andrew},
  title     = {Out of Time: Automated Lip Sync in the Wild},
  booktitle = {Workshop on Multi-view Lip-reading, ACCV},
  year      = {2016}
}

@inproceedings{heusel2017gans,
  author    = {Heusel, Martin and Ramsauer, Hubert and Unterthiner, Thomas and Nessler, Bernhard and Hochreiter, Sepp},
  title     = {{GANs} Trained by a Two Time-Scale Update Rule Converge to a Local {Nash} Equilibrium},
  booktitle = {Advances in Neural Information Processing Systems (NeurIPS)},
  volume    = {30},
  year      = {2017}
}

@inproceedings{unterthiner2019towards,
  author    = {Unterthiner, Thomas and van Steenkiste, Sjoerd and Kurach, Karol and Marinier, Rapha{\"e}l and Michalski, Marcin and Gelly, Sylvain},
  title     = {Towards Accurate Generative Models of Video: {A} New Metric and Challenges},
  booktitle = {International Conference on Learning Representations (ICLR)},
  year      = {2019}
}

@misc{zheng2025vbench20,
  title         = {{VBench}-2.0: Advancing Video Generation Benchmark Suite for Intrinsic Faithfulness},
  author        = {Zheng, Dian and Huang, Ziqi and Liu, Hongbo and Zou, Kai and He, Yinan and Zhang, Fan and Gu, Lulu and Zhang, Yuanhan and He, Jingwen and Zheng, Wei-Shi and Qiao, Yu and Liu, Ziwei},
  year          = {2025},
  eprint        = {2503.21755},
  archivePrefix = {arXiv},
  primaryClass  = {cs.CV},
  url           = {https://arxiv.org/abs/2503.21755}
}

@misc{yin2024onestepdiffusiondistributionmatching,
      title={One-step Diffusion with Distribution Matching Distillation}, 
      author={Tianwei Yin and Michaël Gharbi and Richard Zhang and Eli Shechtman and Fredo Durand and William T. Freeman and Taesung Park},
      year={2024},
      eprint={2311.18828},
      archivePrefix={arXiv},
      primaryClass={cs.CV},
      url={https://arxiv.org/abs/2311.18828}, 
}

@misc{tian2024emo,
      title={EMO: Emote Portrait Alive - Generating Expressive Portrait Videos with Audio2Video Diffusion Model under Weak Conditions}, 
      author={Linrui Tian and Qi Wang and Bang Zhang and Liefeng Bo},
      year={2024},
      eprint={2402.17485},
      archivePrefix={arXiv},
      primaryClass={cs.CV}
}

@misc{wans2v,
      title={Wan-S2V: Audio-Driven Cinematic Video Generation}, 
      author={Xin Gao and Li Hu and Siqi Hu and Mingyang Huang and Chaonan Ji and Dechao Meng and Jinwei Qi and Penchong Qiao and Zhen Shen and Yafei Song and Ke Sun and Linrui Tian and Guangyuan Wang and Qi Wang and Zhongjian Wang and Jiayu Xiao and Sheng Xu and Bang Zhang and Peng Zhang and Xindi Zhang and Zhe Zhang and Jingren Zhou and Lian Zhuo},
      year={2025},
      eprint={2508.18621},
      archivePrefix={arXiv},
      primaryClass={cs.CV},
      url={https://arxiv.org/abs/2508.18621}, 
}

@misc{wu2025qwenimagetechnicalreport,
      title={Qwen-Image Technical Report}, 
      author={Chenfei Wu and Jiahao Li and Jingren Zhou and Junyang Lin and Kaiyuan Gao and Kun Yan and Sheng-ming Yin and Shuai Bai and Xiao Xu and Yilei Chen and Yuxiang Chen and Zecheng Tang and Zekai Zhang and Zhengyi Wang and An Yang and Bowen Yu and Chen Cheng and Dayiheng Liu and Deqing Li and Hang Zhang and Hao Meng and Hu Wei and Jingyuan Ni and Kai Chen and Kuan Cao and Liang Peng and Lin Qu and Minggang Wu and Peng Wang and Shuting Yu and Tingkun Wen and Wensen Feng and Xiaoxiao Xu and Yi Wang and Yichang Zhang and Yongqiang Zhu and Yujia Wu and Yuxuan Cai and Zenan Liu},
      year={2025},
      eprint={2508.02324},
      archivePrefix={arXiv},
      primaryClass={cs.CV},
      url={https://arxiv.org/abs/2508.02324}, 
}

@article{liu2025rolling,
  title={Rolling Forcing: Autoregressive Long Video Diffusion in Real Time},
  author={Liu, Kunhao and Hu, Wenbo and Xu, Jiale and Shan, Ying and Lu, Shijian},
  journal={arXiv preprint arXiv:2509.25161},
  year={2025}
}

@misc{huang2025selfforcingbridgingtraintest,
      title={Self Forcing: Bridging the Train-Test Gap in Autoregressive Video Diffusion}, 
      author={Xun Huang and Zhengqi Li and Guande He and Mingyuan Zhou and Eli Shechtman},
      year={2025},
      eprint={2506.08009},
      archivePrefix={arXiv},
      primaryClass={cs.CV},
      url={https://arxiv.org/abs/2506.08009}, 
}

@misc{cui2025selfforcingminutescalehighqualityvideo,
      title={Self-Forcing++: Towards Minute-Scale High-Quality Video Generation}, 
      author={Justin Cui and Jie Wu and Ming Li and Tao Yang and Xiaojie Li and Rui Wang and Andrew Bai and Yuanhao Ban and Cho-Jui Hsieh},
      year={2025},
      eprint={2510.02283},
      archivePrefix={arXiv},
      primaryClass={cs.CV},
      url={https://arxiv.org/abs/2510.02283}, 
}

@misc{liu2025decoupleddmdcfgaugmentation,
      title={Decoupled DMD: CFG Augmentation as the Spear, Distribution Matching as the Shield}, 
      author={Dongyang Liu and Peng Gao and David Liu and Ruoyi Du and Zhen Li and Qilong Wu and Xin Jin and Sihan Cao and Shifeng Zhang and Hongsheng Li and Steven Hoi},
      year={2025},
      eprint={2511.22677},
      archivePrefix={arXiv},
      primaryClass={cs.CV},
      url={https://arxiv.org/abs/2511.22677}, 
}

@misc{nie2026transitionmatchingdistillationfast,
      title={Transition Matching Distillation for Fast Video Generation}, 
      author={Weili Nie and Julius Berner and Nanye Ma and Chao Liu and Saining Xie and Arash Vahdat},
      year={2026},
      eprint={2601.09881},
      archivePrefix={arXiv},
      primaryClass={cs.CV},
      url={https://arxiv.org/abs/2601.09881}, 
}

@misc{chen2024diffusionforcingnexttokenprediction,
      title={Diffusion Forcing: Next-token Prediction Meets Full-Sequence Diffusion}, 
      author={Boyuan Chen and Diego Marti Monso and Yilun Du and Max Simchowitz and Russ Tedrake and Vincent Sitzmann},
      year={2024},
      eprint={2407.01392},
      archivePrefix={arXiv},
      primaryClass={cs.LG},
      url={https://arxiv.org/abs/2407.01392}, 
}

@misc{cosyvoice,
      title={CosyVoice: A Scalable Multilingual Zero-shot Text-to-speech Synthesizer based on Supervised Semantic Tokens}, 
      author={Zhihao Du and Qian Chen and Shiliang Zhang and Kai Hu and Heng Lu and Yexin Yang and Hangrui Hu and Siqi Zheng and Yue Gu and Ziyang Ma and Zhifu Gao and Zhijie Yan},
      year={2024},
      eprint={2407.05407},
      archivePrefix={arXiv},
      primaryClass={cs.SD},
      url={https://arxiv.org/abs/2407.05407}, 
}

@misc{seedance2026seedance20advancingvideo,
      title={Seedance 2.0: Advancing Video Generation for World Complexity}, 
      author={Team Seedance and De Chen and Liyang Chen and Xin Chen and Ying Chen and Zhuo Chen and Zhuowei Chen and Feng Cheng and Tianheng Cheng and Yufeng Cheng and Mojie Chi and Xuyan Chi and Jian Cong and Qinpeng Cui and Fei Ding and Qide Dong and Yujiao Du and Haojie Duanmu and Junliang Fan and Jiarui Fang and Jing Fang and Zetao Fang and Chengjian Feng and Yu Gao and Diandian Gu and Dong Guo and Hanzhong Guo and Qiushan Guo and Boyang Hao and Hongxiang Hao and Haoxun He and Jiaao He and Qian He and Tuyen Hoang and Heng Hu and Ruoqing Hu and Yuxiang Hu and Jiancheng Huang and Weilin Huang and Zhaoyang Huang and Zhongyi Huang and Jishuo Jin and Ming Jing and Ashley Kim and Shanshan Lao and Yichong Leng and Bingchuan Li and Gen Li and Haifeng Li and Huixia Li and Jiashi Li and Ming Li and Xiaojie Li and Xingxing Li and Yameng Li and Yiying Li and Yu Li and Yueyan Li and Chao Liang and Han Liang and Jianzhong Liang and Ying Liang and Wang Liao and J. H. Lien and Shanchuan Lin and Xi Lin and Feng Ling and Yue Ling and Fangfang Liu and Jiawei Liu and Jihao Liu and Jingtuo Liu and Shu Liu and Sichao Liu and Wei Liu and Xue Liu and Zuxi Liu and Ruijie Lu and Lecheng Lyu and Jingting Ma and Tianxiang Ma and Xiaonan Nie and Jingzhe Ning and Junjie Pan and Xitong Pan and Ronggui Peng and Xueqiong Qu and Yuxi Ren and Yuchen Shen and Guang Shi and Lei Shi and Yinglong Song and Fan Sun and Li Sun and Renfei Sun and Wenjing Tang and Boyang Tao and Zirui Tao and Dongliang Wang and Feng Wang and Hulin Wang and Ke Wang and Qingyi Wang and Rui Wang and Shuai Wang and Shulei Wang and Weichen Wang and Xuanda Wang and Yanhui Wang and Yue Wang and Yuping Wang and Yuxuan Wang and Zijie Wang and Ziyu Wang and Guoqiang Wei and Meng Wei and Di Wu and Guohong Wu and Hanjie Wu and Huachao Wu and Jian Wu and Jie Wu and Ruolan Wu and Shaojin Wu and Xiaohu Wu and Xinglong Wu and Yonghui Wu and Ruiqi Xia and Xin Xia and Xuefeng Xiao and Shuang Xu and Bangbang Yang and Jiaqi Yang and Runkai Yang and Tao Yang and Yihang Yang and Zhixian Yang and Ziyan Yang and Fulong Ye and Bingqian Yi and Xing Yin and Yongbin You and Linxiao Yuan and Weihong Zeng and Xuejiao Zeng and Yan Zeng and Siyu Zhai and Zhonghua Zhai and Bowen Zhang and Chenlin Zhang and Heng Zhang and Jun Zhang and Manlin Zhang and Peiyuan Zhang and Shuo Zhang and Xiaohe Zhang and Xiaoying Zhang and Xinyan Zhang and Xinyi Zhang and Yichi Zhang and Zixiang Zhang and Haiyu Zhao and Huating Zhao and Liming Zhao and Yian Zhao and Guangcong Zheng and Jianbin Zheng and Xiaozheng Zheng and Zerong Zheng and Kuan Zhu and Feilong Zuo},
      year={2026},
      eprint={2604.14148},
      archivePrefix={arXiv},
      primaryClass={cs.CV},
      url={https://arxiv.org/abs/2604.14148}, 
}

@misc{jiang2025omnihuman15instillingactivemind,
      title={OmniHuman-1.5: Instilling an Active Mind in Avatars via Cognitive Simulation}, 
      author={Jianwen Jiang and Weihong Zeng and Zerong Zheng and Jiaqi Yang and Chao Liang and Wang Liao and Han Liang and Yuan Zhang and Mingyuan Gao},
      year={2025},
      eprint={2508.19209},
      archivePrefix={arXiv},
      primaryClass={cs.CV},
      url={https://arxiv.org/abs/2508.19209}, 
}

@inproceedings{liu2025javisdit,
  title       = {JavisDiT: Joint Audio-Video Diffusion Transformer with Hierarchical Spatio-Temporal Prior Synchronization}, 
  author      = {Liu, Kai and Li, Wei and Chen, Lai and Wu, Shengqiong and Zheng, Yanhao and Ji, Jiayi and Zhou, Fan and Luo, Jiebo and Liu, Ziwei and Fei, Hao and Chua, Tat-Seng},
  conference  = {The Fourteenth International Conference on Learning Representations},
  year        = {2026},
}

@misc{peng2025vibevoicetechnicalreport,
      title={VibeVoice Technical Report}, 
      author={Zhiliang Peng and Jianwei Yu and Wenhui Wang and Yaoyao Chang and Yutao Sun and Li Dong and Yi Zhu and Weijiang Xu and Hangbo Bao and Zehua Wang and Shaohan Huang and Yan Xia and Furu Wei},
      year={2025},
      eprint={2508.19205},
      archivePrefix={arXiv},
      primaryClass={cs.CL},
      url={https://arxiv.org/abs/2508.19205}, 
}

@misc{li2026halloliverealtimestreamingjoint,
      title={Hallo-Live: Real-Time Streaming Joint Audio-Video Avatar Generation with Asynchronous Dual-Stream and Human-Centric Preference Distillation}, 
      author={Chunyu Li and Jiaye Li and Ruiqiao Mei and Haoyuan Xia and Hao Zhu and Jingdong Wang and Siyu Zhu},
      year={2026},
      eprint={2604.23632},
      archivePrefix={arXiv},
      primaryClass={cs.CV},
      url={https://arxiv.org/abs/2604.23632}, 
}

@misc{openmoss_mova_2026,
  title         = {MOVA: Towards Scalable and Synchronized Video-Audio Generation},
  author        = {{SII-OpenMOSS Team} and Donghua Yu and Mingshu Chen and Qi Chen and Qi Luo and Qianyi Wu and Qinyuan Cheng and Ruixiao Li and Tianyi Liang and Wenbo Zhang and Wenming Tu and Xiangyu Peng and Yang Gao and Yanru Huo and Ying Zhu and Yinze Luo and Yiyang Zhang and Yuerong Song and Zhe Xu and Zhiyu Zhang and Chenchen Yang and Cheng Chang and Chushu Zhou and Hanfu Chen and Hongnan Ma and Jiaxi Li and Jingqi Tong and Junxi Liu and Ke Chen and Shimin Li and Songlin Wang and Wei Jiang and Zhaoye Fei and Zhiyuan Ning and Chunguo Li and Chenhui Li and Ziwei He and Zengfeng Huang and Xie Chen and Xipeng Qiu},
  year          = {2026},
  month         = feb,
  eprint        = {2602.08794},
  archivePrefix = {arXiv},
  primaryClass  = {cs.CV},
  doi           = {10.48550/arXiv.2602.08794},
  url           = {https://arxiv.org/abs/2602.08794},
  note          = {Technical report. Corresponding authors: Xie Chen and Xipeng Qiu. Project leaders: Qinyuan Cheng and Tianyi Liang.}
}

@misc{zeng2026lpm10videobasedcharacter,
      title={LPM 1.0: Video-based Character Performance Model}, 
      author={Ailing Zeng and Casper Yang and Chauncey Ge and Eddie Zhang and Garvey Xu and Gavin Lin and Gilbert Gu and Jeremy Pi and Leo Li and Mingyi Shi and Shawn Wang and Sheng Bi and Steven Tang and Thorn Hang and Tobey Guo and Vincent Li and Xin Tong and Yikang Li and Yuchen Sun and Yue Zhao and Yuhan Lu and Yuwei Li and Zane Zhang and Zeshi Yang and Zi Ye},
      year={2026},
      eprint={2604.07823},
      archivePrefix={arXiv},
      primaryClass={cs.CV},
      url={https://arxiv.org/abs/2604.07823}, 
}

\clearpage
\pagestyle{empty} 


\end{document}